\documentclass[10pt, conference, letterpaper]{ IEEEtran}
\IEEEoverridecommandlockouts
\usepackage{cite}
\usepackage{amsmath,amssymb,amsfonts,amsthm}
\usepackage{algorithmic}
\usepackage{graphicx}
\usepackage{textcomp}
\usepackage{xcolor}
\def\BibTeX{{\rm B\kern-.05em{\sc i\kern-.025em b}\kern-.08em
    T\kern-.1667em\lower.7ex\hbox{E}\kern-.125emX}}

\theoremstyle{definition}
\newtheorem{assumption}{Assumption}
\newtheorem{theorem}{Theorem}
\newtheorem{definition}{Definition}

\newtheorem{lemma}{Lemma}
\newtheorem{remark}{Remark}

\usepackage[hidelinks]{hyperref}
\usepackage{enumitem}
\usepackage{multirow}
\usepackage{subcaption}
\usepackage[linesnumbered,ruled,vlined]{algorithm2e}
\allowdisplaybreaks[1]

\usepackage{todonotes}

\renewcommand{\Pr}{\mathbb{P}}
\usepackage{booktabs}
\begin{document}

\title{Certified Unlearning in Decentralized Federated Learning}
\author{\IEEEauthorblockN{
Hengliang Wu\IEEEauthorrefmark{2}\IEEEauthorrefmark{1},
Youming Tao\IEEEauthorrefmark{2}\IEEEauthorrefmark{3}\IEEEauthorrefmark{1},
Anhao Zhou\IEEEauthorrefmark{2},
Shuzhen Chen\IEEEauthorrefmark{4},
Falko Dressler\IEEEauthorrefmark{3},
Dongxiao Yu\IEEEauthorrefmark{2}
}
\IEEEauthorblockA{\IEEEauthorrefmark{2}School of Computer Science and Technology, Shandong University, Qingdao, China\\
\IEEEauthorrefmark{3}School of Electrical Engineering and Computer Science, TU Berlin, Berlin, Germany\\
\IEEEauthorrefmark{4}College of Computer Science and
Technology, Ocean University of China, Qingdao, China}

\thanks{\IEEEauthorrefmark{1}The first two authors contributed equally to this work.}
\thanks{Corresponding Author: Shuzhen Chen (Email: szchen@ouc.edu.cn).}
}







\maketitle

\begin{abstract}
Driven by the right to be forgotten (RTBF), machine unlearning has become an essential requirement for privacy-preserving machine learning. However, its realization in decentralized federated learning (DFL) remains largely unexplored. In DFL, clients exchange local updates only with neighbors, causing model information to propagate and mix across the network. As a result, when a client requests data deletion, its influence is implicitly embedded throughout the system, making removal difficult without centralized coordination. We propose a novel certified unlearning framework for DFL based on Newton-style updates. Our approach first quantifies how a client's data influence propagates during training. Leveraging curvature information of the loss with respect to the target data, we then construct corrective updates using Newton-style approximations. To ensure scalability, we approximate second-order information via Fisher information matrices. The resulting updates are perturbed with calibrated noise and broadcast through the network to eliminate residual influence across clients. We theoretically prove that our approach satisfies the formal definition of certified unlearning, ensuring that the unlearned model is difficult to distinguish from a retrained model without the deleted data. We also establish utility bounds showing that the unlearned model remains close to retraining from scratch. Extensive experiments across diverse decentralized settings demonstrate the effectiveness and efficiency of our framework.
\end{abstract}

\begin{IEEEkeywords}
Machine Unlearning, Federated Learning, Decentralized Learning 
\end{IEEEkeywords}

\section{Introduction}
Federated Learning (FL) enables collaborative model training while keeping user data on local devices, sharing only model updates among participants~\cite{fedavg,chen2025adaptive}, which offers a promising approach to privacy-preserving learning. However, conventional FL relies on a centralized server to aggregate updates, which introduces several limitations, including communication bottlenecks, heavy server load, and vulnerability to single-point failures.
To overcome these issues, \emph{Decentralized Federated Learning} (DFL) has emerged as an attractive alternative~\cite{dpsgd,tao2023distributed}. In DFL, clients exchange model updates directly with neighboring peers rather than communicating with a central coordinator. This peer-to-peer design distributes communication costs across the network, improves scalability and fault tolerance, and eliminates reliance on a single trusted server. As a result, DFL has attracted increasing attention in both academic research and real-world deployments, such as large-scale distributed and vehicular systems~\cite{DFLsurvey, vehicularDPSGD}.
\par 
The privacy guarantees of FL and DFL are increasingly challenged. Prior work has demonstrated that sensitive user data can be extracted from trained models through inference and reconstruction attacks~\cite{Userlevelprivacyleakage}. In response, governments have enacted stricter data protection regulations, most notably the \emph{right to be forgotten} (RTBF)~\cite{gdpr, legistion}, which grants users the right to request the deletion of their sensitive data from any models derived from it. A na\"ive solution is to retrain the model from scratch after removing the targeted data, but this approach is often prohibitively expensive. This challenge has motivated the development of \emph{machine unlearning}, which aims to efficiently remove the influence of specific data points or clients without full retraining~\cite{unlearning}.
\par
To meet RTBF requirements in federated settings, \emph{Federated Unlearning} (FU) has recently emerged~\cite{FUsurvey}. Existing FU methods primarily focus on centralized FL architectures~\cite{federaser, Fedrecovery, FAT}. However, extending unlearning to decentralized environments, referred to as \emph{Decentralized Federated Unlearning} (DFU), remains largely unexplored. Unlearning in decentralized settings is inherently difficult due to the following challenges.

\par\noindent\textbf{(1) Implicit influence propagation.} Although communication occurs only among neighboring clients, each local model implicitly incorporates gradient information from all participants through repeated mixing. When a client requests data deletion, its contribution has already diffused throughout the network. Precisely quantifying and removing this mixed influence without centralized coordination is non-trivial.

\par\noindent\textbf{(2) Efficiency constraints.}
DFU must be computationally and storage efficient, as decentralized learning often involves resource-constrained edge devices. Existing DFU approaches frequently rely on storing historical updates or auxiliary seed models to support unlearning~\cite{hdus, PDUDT}, resulting in substantial memory and computation overhead. This motivates the need for unlearning algorithms specifically designed for decentralized environments with minimal storage and computation requirements.

\par\noindent\textbf{(3) Certified guarantees and utility preservation.}
A principled unlearning algorithm must satisfy \emph{certified unlearning} guarantees, ensuring that the unlearned model is difficult to distinguish from one trained without the deleted data~\cite{RememberWhatYouWanttoForget, certified}. At the same time, it must preserve model utility, which requires tight bounds on the deviation between the unlearned model and the retrained model. Existing DFU methods either lack formal unlearning guarantees~\cite{hdus} or fail to rigorously characterize the resulting utility loss~\cite{PDUDT}.

\par\noindent\textbf{Our Contributions.}
To address these challenges, we propose a generic and efficient framework for \emph{certified decentralized federated unlearning} applicable to arbitrary network topologies. Our key contributions are summarized as follows:

\begin{itemize}
\item We propose a novel DFU framework that achieves certified unlearning in decentralized learning. Our approach quantifies peer-to-peer influence using the doubly stochastic property of network mixing, constructs corrective updates via Newton-style approximations, and injects calibrated noise to ensure indistinguishability. To improve efficiency, we approximate second-order information using the Fisher Information Matrix (FIM).
\item We provide rigorous theoretical guarantees for our method. Specifically, we prove that our algorithm satisfies the formal definition of $(\epsilon,\delta)$-certified unlearning and establish error bounds between the unlearned model and a fully retrained model, ensuring utility preservation. To the best of our knowledge, this is the first certified DFU framework that does not require storing historical updates.
\item We conduct extensive experiments across diverse decentralized settings. The results demonstrate that our method achieves accuracy comparable to retraining from scratch while reducing time cost by up to 97\%, highlighting both its effectiveness and efficiency.
\end{itemize}
\section{Related Work}
\textbf{Machine unlearning.} Machine unlearning was first formalized by Cao et al.~\cite{systemsforget}, who defined unlearning as producing a model equivalent to one obtained by retraining after removing the target data. Subsequent work has proposed alternative definitions, broadly categorized into \emph{exact} and \emph{approximate} unlearning~\cite{MUsurvey}. Exact unlearning requires that the distribution of the unlearned model be identical to that of a model retrained from scratch on the retained data~\cite{rtbf-in-fl}, thereby achieving statistical indistinguishability. Due to its stringent requirements, exact unlearning is often computationally infeasible. To address this limitation, approximate unlearning, most notably $(\epsilon,\delta)$-unlearning, has been widely adopted~\cite{RememberWhatYouWanttoForget, certified}. Under this definition, the output distribution of the unlearning algorithm is \emph{difficult to distinguish} from that of retraining, with the distinguishing advantage bounded by $(\epsilon,\delta)$. While approximate unlearning relaxes the requirements of exact unlearning and significantly reduces computation and memory costs, existing methods are primarily designed for centralized systems and do not extend to decentralized learning settings~\cite{certified, hessianfreeUn, AdversarialMU}.
\par
\textbf{Federated unlearning.} With the rise of federated learning, machine unlearning has been extended to federated settings, giving rise to \emph{Federated Unlearning} (FU)~\cite{FUsurvey}. FU methods typically consider three levels of granularity: sample-wise, class-wise, and client-wise unlearning~\cite{NoT, Reversible-federated-unlearning}. Sample-wise unlearning removes individual data points, class-wise unlearning deletes all samples of a specific class, and client-wise unlearning removes all contributions from a particular client~\cite{Reversible-federated-unlearning}. Most existing FU algorithms assume a centralized FL architecture~\cite{FAT}. Prior work explores techniques such as momentum degradation~\cite{F-Un_momentum}, model perturbation~\cite{SIfu}, and other heuristic strategies~\cite{FUsurvey}. In contrast, our approach is grounded in Newton-style updates, offering a principled and intuitive mechanism for influence removal. Moreover, it is compatible with a broad class of Empirical Risk Minimization (ERM) objectives, enabling strong generalization performance. 
\par 
\textbf{Decentralized unlearning.} Research on unlearning in fully decentralized federated learning remains limited. Ye et al.~\cite{hdus} proposed an exact decentralized unlearning method based on model distillation. While empirical results suggest effectiveness, the absence of formal unlearning guarantees limits its reliability. More recently, Qiao et al.~\cite{PDUDT} introduced the first provable client-wise decentralized unlearning algorithm. Their method estimates client influence using aggregated historical gradients and applies perturbations to achieve unlearning. However, this approach requires storing historical gradient information and performing multiple additional training rounds to recover accuracy, leading to substantial storage and computational overhead. These limitations hinder its applicability in resource-constrained decentralized environments. In contrast, our work aims to design an efficient and certified decentralized unlearning framework that avoids historical update storage while providing rigorous unlearning and utility guarantees.

\section{Problem Formulation}
\subsection{Decentralized Learning} 
We consider a fully decentralized FL system with $N$ clients, where clients communicate exclusively with their neighbors and no central server is present. Each client $i$ ($i\in[N]$) holds a local dataset $S_i$ containing $n_i$ samples drawn from a client-specific local distribution $\mathcal{D}_i$.  The communication structure of the system is captured by a graph $(V,E,Q)$. Here, $V$ is the set of $N$ nodes representing clients, and $E$ denotes the set of communication links between clients. The matrix $Q\in\mathbb{R}^{N \times N}$ is a mixing matrix that characterizes how model updates are weighted when exchanging among clients. Specifically, the entry $Q_{ij}$ quantifies the influence of client-$j$ on client $i$ during local aggregation.
We assume that $Q$ is a symmetric doubly stochastic matrix, i.e., $Q_{ij}=Q_{ji}$ for all $i,j$ and $\sum_{i}Q_{ji}=1$ for all $j$. If clients $i$ and $j$ are not connected, $Q_{ij}=0$. This ensures that communication is restricted to neighbouring clients and that the aggregation process preserves consensus properties. The matrix $Q$ further satisfies the following standard spectral assumption.
\begin{assumption}\label{assumptionQ}
Given the symmetric doubly stochastic matrix $Q$, and let $\rho_k(Q)$ denote the $k$-th largest eigenvalue of $Q$. Define $\rho:=(\max\{|\rho_2(Q)|,|\rho_N(Q)|\})^2$. We assume $\rho<1$.
\end{assumption}
\par The learning objective is to solve the following decentralized optimization problem:
\begin{equation}
    \min_{x\in\mathbb{R}^d}f(x)=\frac{1}{N}\sum_{i=1}^{N}f_i(x),\label{eq1}
\end{equation} 
where $f_i(x)$ is the local empirical risk function defined as $f_i(x)=\frac{1}{n_i}\sum_{t=1}^{n_i}F_i(x,\xi_{i,t})$ and $F_i(x,\xi_{i,t})$ is the loss incurred by model parameter $x$ on data sample $\xi_{i,t}\in S_i$. During training, clients collaboratively optimize~\eqref{eq1} through local updates and communication. After total $K$ iterations, each client maintains a local model $x_{K,i}$, and the final global model is given by the average $\bar{x}=\frac{1}{N}\sum_{i=1}^Nx_{K,i}$, which is assumed to converge to a stationary solution of~\eqref{eq1}.

\subsection{Decentralized Unlearning}
\par 
Let $\mathcal{S}$ denote the data universe and $\mathcal{X}\subseteq\mathbb{R}^d$ denote the model space.
Client $i\in[N]$ holds a dataset $S_i \in \mathcal{S}^{n_i}$, and the collection of all clients' datasets is denoted by
$\mathbf{S}:=(S_1,\dots,S_N)\in\mathcal{S}^{n_1}\times\cdots\times\mathcal{S}^{n_N}$.
A decentralized learning algorithm is a (randomized) mapping $\mathcal{A}:\ \mathcal{S}^{n_1}\times\cdots\times\mathcal{S}^{n_N}\ \to\ \mathcal{X}^N$,
which outputs a tuple of local models $\mathbf{X}:=(x_1,\dots,x_N)\in\mathcal{X}^N$. For a deletion request issued by client $i$, let $U_i\subseteq S_i$ with $|U_i|=m_i$ and define the retained dataset
$R_i:=S_i\setminus U_i$.
We use $\mathbf{S}^{(-i)}:=(S_1,\dots,R_i,\dots,S_N)$ to denote the dataset collection after deleting $U_i$ from client $i$.
Let $\mathcal{T}$ denote the space of auxiliary information available to the unlearning procedure (e.g., statistics computed during training).
A decentralized unlearning algorithm is a (possibly randomized) mapping
$\mathcal{G}:\mathcal{P}(\mathcal{S})\times \mathcal{X}^N \times \mathcal{T} \to\ \mathcal{X}^N$,
which takes as input the deletion request $U_i$, the learned models $\mathcal{A}(\mathbf{S})$, and auxiliary information $T(\mathbf{S})\in\mathcal{T}$,
and outputs the unlearned models $\widetilde{\mathbf{X}}:=(\widetilde{x}_1,\dots,\widetilde{x}_N)\in\mathcal{X}^N$.

We extend the notion of certified $(\epsilon,\delta)$-unlearning~\cite{RememberWhatYouWanttoForget} to the decentralized setting, formalized as follows. 

\begin{definition}[Decentralized $(\epsilon,\delta)$-unlearning]\label{defUn}
We say that the pair $(\mathcal{A},\mathcal{G})$ satisfies \emph{certified $(\epsilon,\delta)$-unlearning} in decentralized learning if, for any measurable set
$W\subseteq\mathcal{X}^N$,
\begin{align*}
&\Pr\!\left(
\mathcal{G}\!\left(U_i,\mathcal{A}(\mathbf{S}),T(\mathbf{S})\right)\in W
\right)
\le
e^{\epsilon}
\Pr\!\left(
\mathcal{A}(\mathbf{S}^{(-i)})\in W
\right)
+\delta, \\
&\Pr\!\left(
\mathcal{A}(\mathbf{S}^{(-i)})\in W
\right)
\le
e^{\epsilon}
\Pr\!\left(
\mathcal{G}\!\left(U_i,\mathcal{A}(\mathbf{S}),T(\mathbf{S})\right)\in W
\right)
+\delta.
\end{align*}
\end{definition}


\section{Certified Decentralized Unlearning Framework}
\subsection{Decentralized Learning Process}
\par During training, each client $i$ collaboratively optimizes the global objective in a fully decentralized manner. The learning process proceeds for $K$ iterations and follows standard decentralized stochastic gradient descent (DSGD) scheme. At each iteration $k\in[K]$, the training proceeds as the following steps.

\par\noindent\textbf{Step 1 (Local stochastic gradient update).} Client $i$ samples a data point $\xi_{k,i}$ uniformly at random from its local dataset $S_i$ and computes the stochastic gradient $\nabla F_i(x_{k,i},\xi_{k,i})$ based on its current local model $x_{k,i}$.
\par\noindent\textbf{Step 2 (Model exchange and aggregation).}
Each client exchanges its local model $x_{k,i}$ with its neighbours and aggregates the received models using the mixing $Q$. The aggregated model is given by $x_{k+\frac{1}{2},i}=\sum_jQ_{ij}x_{k,j}$.
\par\noindent\textbf{Step 3 (Local model update).}
Client $i$ updates its local model using a gradient descent step based on the aggregated model: $x_{k+1,i}=x_{k+\frac{1}{2},i}-\gamma\nabla F_i(x_{k,i},\xi_{k,i})$, where $\gamma>0$ denotes the learning rate.
\par\noindent\textbf{Step 4 (Termination and statistic collection).}
After $K$ iterations, each client $i$ obtains a local model $x_{K,i}$ that approximates a stationary solution of~\eqref{eq1}, with convergence guarantees under standard assumptions, see e.g.,~\cite{dpsgd, DSGDconvergence}. If a deletion request is issued after iteration $k$, each client $i$ additionally computes and stores the auxiliary statistic $T_i(S_i):=\{\nabla^2 f_i(x_{k,i})\}$, i.e., the Hessian of the local empirical risk evaluated at the model state corresponding to the request time.
The auxiliary statistic $T_i(S_i)$ is used exclusively for certified unlearning. Let $d$ be the model dimension. Importantly, storing $T_i(S_i)$ requires $\mathcal{O}(d^2)$ memory , which is independent of both the local dataset size $n_i$ and the number of deleted samples $m_i$, making it scalable and suitable for resource-constrained decentralized environments.

\subsection{Decentralized Unlearning for Sample-Level Removal}
We start with \emph{sample-wise certified decentralized unlearning}, where a client requests to delete a subset of its local data.

\textbf{Influence propagation in decentralized learning.}
Due to iterative peer-to-peer communication, the contribution of each client is gradually mixed into all local models.
As a result, removing the effect of deleted samples requires accounting for how model updates propagate through the network.
While prior work~\cite{PDUDT} estimates such influence by tracking historical gradients, this approach incurs substantial storage and computation overhead.
In contrast, we exploit the structural properties of the mixing matrix $Q$.
When $Q$ is symmetric and doubly stochastic, repeated local averaging leads to asymptotic consensus among clients.
Specifically, peer-to-peer influence weights converge to the uniform distribution $1/N$ as the number of iterations increases~\cite{dpsgd}.
This behavior is formalized in Lemma~\ref{lemmaWeight}, which states that the peer-to-peer influence weights converge exponentially fast to uniform consensus. This allows us to approximate the influence of each client on any other client after $k$ iterations using a uniform weighting scheme, eliminating the need to store historical gradient information and significantly reducing memory and computational costs.
\begin{lemma}\label{lemmaWeight}
Under Assumption~\ref{assumptionQ}, the mixing matrix $Q$ satisfies
$\left\| Q^k - \frac{1}{N}\mathbf{1}_N \mathbf{1}_N^\top \right\|^2\le\rho^k$, $\forall, k \in \mathbb{N}$,
where $\rho<1$ is defined in Assumption~\ref{assumptionQ} and $\mathbf{1}_N\in\mathbb{R}^N$ denotes the all-ones vector.
\end{lemma}
\textbf{Newton-style correction via influence functions.}
Consider a client $c \in [N]$ that issues a deletion request $U_c \subseteq S_c$ at iteration $k$, where $|U_c| = m_c$.
To estimate the impact of removing these samples, we draw inspiration from influence functions~\cite{influence-functions} and approximate the parameter change induced by deleting $U_c$ using a second-order expansion of the local empirical risk around the current model $x_{k,c}$. We made the following standard assumptions on loss functions.
\begin{assumption}\label{assumptionStrong-convex}
    Loss functions $F_i(x,\xi)$ are $\lambda$-strongly convex, $\mu$-smooth, $L$-Lipschitz with $M$-Lipschitz Hessian.
\end{assumption}
We define the empirical Hessian over the retained data as
\begin{equation}
    \hat{H_c}=\frac{1}{n_c-m_c} {\textstyle \sum_{\xi_c\in S_c\setminus U_c}}\nabla ^2F_c(x_{k,c},\xi_c ).
\end{equation}
Using this curvature information, we construct a Newton-style corrective update, which approximates the change in model parameters caused by deleting the samples in $U_c$.
\begin{equation}
    x_c^{\Delta} =\frac{1}{n_c-m_c}(\hat{H_c})^{-1} {\textstyle \sum_{\xi_c\in U_c}}\nabla F_c(x_{k,c},\xi_c ).\label{x_delta}
\end{equation}
Notably, this approximation relies only on second-order information of the loss landscape and gradient evaluations at the deleted samples, without requiring access to raw data beyond what is already locally available.

\textbf{Perturbation and decentralized aggregation.}
To ensure certified unlearning guarantees, client $c$ perturbs the corrective update by adding Gaussian noise
$\nu_c \sim \mathcal{N}(0, \sigma_c^2 I_d)$,
where the noise variance $\sigma_c^2$ is carefully calibrated based on the sensitivity of the update.
The perturbed update is then broadcast through the network. Leveraging the uniform influence approximation, all clients aggregate the received corrective updates using a uniform weight $1/N$ and apply the aggregated correction to their local models.
This procedure removes the influence of the deleted samples from all local models without centralized coordination.
The complete process for sample-level removal is summarized in Algorithm~\ref{alg}.

\begin{remark}[Optional fine-tuning for utility recovery]
Following unlearning, a lightweight fine-tuning step may be applied to further improve model utility, as adopted in prior work~\cite{NoT, PDUDT}.
This step follows the standard D-PSGD procedure and is performed exclusively on the retained data.
Our experiments demonstrate that a \emph{single} fine-tuning round is sufficient to match the performance of a fully retrained model. In contrast to prior approaches that rely on many post-unlearning fine-tuning rounds, e.g. up to 200 rounds in~\cite{PDUDT}, our method yields significantly lower computation and communication overheads.
Moreover, we prove that this step preserves the $(\epsilon,\delta)$-certified unlearning guarantee, see Section~\ref{sec5}.
\end{remark}

\begin{remark}[Multiple simultaneous deletion requests]
The proposed framework naturally supports multiple clients $\mathcal{C}\subseteq[N]$ issuing deletion requests simultaneously.
To preserve utility guarantees, the total number of deleted samples must satisfy
\(
\sum_{c\in\mathcal{C}} m_c \le \zeta,
\)
where $\zeta$ denotes the deletion capacity.
The same upper and lower bounds on deletion capacity established in~\cite{RememberWhatYouWanttoForget} apply to our framework.
\end{remark}


\begin{remark}[Extension to general convex losses]
Although our analysis assumes strong convexity, the proposed algorithm can be extended to general convex loss functions.
Specifically, for a convex loss function $F_i(\cdot,\xi)$, we define the regularized loss
\begin{equation}
\widetilde{F}_i(x,\xi)
=
F_i(x,\xi)
+
\frac{\lambda}{2}\|x\|^2,
\end{equation}
which is $\lambda$-strongly convex.
Applying our decentralized unlearning algorithm to the regularized objective yields the same form of updates and certified unlearning guarantees.
This standard regularization technique allows our framework to handle general convex objectives without modifying the algorithmic structure.
\end{remark}

\begin{algorithm}[t]
\caption{Certified Hessian-based Decentralized Sample-Wise Unlearning Algorithm}
\label{alg}
\KwIn{
    Client $c \in [C]$ submitting deletion requests ($[C] \subseteq [N]$), Client $i \in [N]$;
    Delete requests: $U_c = \{\xi_{c,t}\}_{t=1}^{m_c} \subseteq S_c$;
    local model: $x_{k,c}$;
    Additional statistic $T_i(S_i): \{\nabla^2 f_i(x_{k,i})\}$;
    Loss function: $F_i(x, \xi)$;
    Set $\Delta F_c = \frac{2ML^{2}m_c^{2}}{\lambda^{3} n_c^{2}}$, $\sigma_c=\frac{\Delta F_c}{\epsilon}\sqrt{2\ln(1.25/\delta )}$
}
\KwOut{$\widetilde{x}_i$ for each client $i$}

\For{\textbf{\textup{every client}} $c\in[C]$ \textbf{\textup{in parallel}}}{
    $\displaystyle
        \hat{H}_c \gets \frac{1}{n_c-m_c} \sum_{\xi_c\in S_c \setminus U_c} \nabla^2 F_c(x_{k,c}, \xi_c)$\;
    $\displaystyle
        x_c^{\Delta} \gets \frac{1}{n_c - m_c} (\hat{H}_c)^{-1} \sum_{\xi_c\in U_c} \nabla F_c(x_{k,c}, \xi_c)$\;
    Sample $\nu_c \in \mathbb{R}^d$ from $\mathcal{N}(0, \sigma_c^2 \mathbb{I}_d)$\;
    $\widetilde{x}_c^{\Delta}\gets x_c^{\Delta}+\nu_c$\;
    Send $\widetilde{x}_c^{\Delta}$ to neighbors\;
    
}

\For{\textbf{\textup{every client}} $i\in[N]$ \textbf{\textup{in parallel}}}{
    $\widetilde{x}_i\gets x_{k,i}$\;
    \If {receiving a message $\widetilde{x}_c^{\Delta}$ for the first time}{
        Forward the message to all other neighbors\; 
        $\displaystyle \widetilde{x}_i \gets \widetilde{x}_i + \frac{1}{N} \widetilde{x}_c^{\Delta}$\;
    }
    \If{receiving a message again}{
        Discard the message\;
    }
}
\Return{$\displaystyle \{\widetilde{x}_i\}_{i=1}^N$}
\end{algorithm}

\subsection{Decentralized Unlearning for Client-Level Removal}

We next extend our framework to \emph{client-wise certified decentralized unlearning}, which aims to completely remove the contribution of a departing client from the trained models.
Unlike sample-wise unlearning, client-wise unlearning requires eliminating the influence of \emph{all} data held by the target client.
Consider a client $c \in [N]$ that requests full removal at iteration $k$.
Following the same influence-function-based philosophy as in the sample-wise setting, we approximate the effect of removing client $c$’s entire dataset using a Hessian-based Newton-style correction. 
To compute the corrective update, client $c$ first collects the auxiliary statistics
$T_i(S_i)=\{\nabla^2 f_i(x_{k,i})\}$ from all remaining clients. Let $f_i(x)=\frac{1}{n_i}\sum_{t=1}^{n_i}F_i(x,\xi_{i,t})$ be the local empirical risk.
Using these statistics, the aggregated Hessian over the retained clients is computed as
\begin{equation}
\hat{H}_c
=
\frac{1}{N-1}
\left(
\sum_{i \in [N]} \nabla^2 f_i(x_{k,i})
-
\nabla^2 f_c(x_{k,c})
\right),
\end{equation}
which approximates the curvature of the global objective after removing client $c$.
Based on the aggregated curvature, the corrective update is given by
\begin{equation}
x_c^{\Delta}
=
\frac{1}{N-1}
(\hat{H}_c)^{-1}
\nabla f_c(x_{k,c}).
\end{equation}
To satisfy certified unlearning guarantees, the corrective update is perturbed with calibrated Gaussian noise and broadcast through the network.
All remaining clients aggregate the received update using uniform weights and apply the correction to their local models, following the same procedure described in Algorithm~\ref{alg}.

\subsection{Efficient Hessian Approximation}
Our unlearning algorithm relies on second-order information through Hessian matrices.
Suppose the model dimension is $d$.
Storing a full Hessian requires $\mathcal{O}(d^2)$ memory, which can still be prohibitive for resource-constrained clients.
To further reduce this overhead, we adopt an efficient approximation based on the Fisher Information Matrix (FIM).
By leveraging the Fisher approximation, the storage consumption can be reduced from $\mathcal{O}(d^2)$ to $\mathcal{O}(d)$, yielding significant memory savings.
In practical settings, the empirical Fisher matrix provides an effective approximation of the Hessian and has been shown to capture its curvature structure even with relatively small sample sizes~\cite{woodfisher}.
Specifically, the empirical Fisher approximation $\Psi_c$ at client $c$ and iteration $k$ is defined as
\begin{equation}
    \Psi_c=\frac{1}{n_c-m_c} \sum_{\xi_c\in S_c\setminus U_c} \nabla F_c(x_{k,c},\xi_c) \nabla F_c(x_{k,c},\xi_c)^T.
\end{equation}
We note that the Fisher approximation is well defined when the loss function corresponds to a negative log-likelihood, i.e.,
$F_i(x,\xi) \propto -\log \mathbb{P}(\xi \mid x)$.
This condition is satisfied by commonly used losses such as squared loss and cross-entropy loss, making the approximation broadly applicable.

Using the Fisher approximation amounts to replacing the empirical Hessian $\hat{H}_c$ with $\Psi_c$ in the unlearning update.
Since $\Psi_c$ is only an approximation of $\hat{H}_c$, certified unlearning guarantees require that the two matrices coincide in certain regimes.
Let $x_c^* = \arg\min_{x \in \mathcal{X}} f_c(x)$ denote the empirical risk minimizer at client $c$.
Prior work~\cite{Fedfisher} shows that $\Psi_c = \hat{H}_c$ at $x_c^*$.
Under this condition, all theoretical guarantees established for Algorithm~\ref{alg} directly extend to the Fisher-based variant without loss of generality.

\subsection{Comparisons to previous work PDUDT\cite{PDUDT}}
We now compare our framework with PDUDT~\cite{PDUDT}, a recent provable approach for decentralized approximate unlearning.

\noindent\textbf{1. Scope and generality.}
PDUDT focuses exclusively on client-wise unlearning, aiming to remove the influence of an entire client.
In contrast, our framework is more general and supports all standard unlearning granularities, including sample-wise, class-wise, and client-wise unlearning, within a unified framework.

\noindent\textbf{2. Utility guarantees.}
Our method provides rigorous utility guarantees: we establish explicit bounds on the test error between the unlearned model and a fully retrained model.
These bounds match state-of-the-art results for certified unlearning~\cite{RememberWhatYouWanttoForget}.
In contrast, the utility degradation induced by PDUDT is difficult to characterize analytically, and no comparable error bounds are provided.

\noindent\textbf{3. Computational and storage efficiency.}
Efficiency is another key distinction.
PDUDT requires an extensive post-unlearning retraining phase consisting of $T$ additional training rounds (e.g., $T=200$ in their experiments), whereas our method requires at most a single fine-tuning round.
We empirically compare the time required to recover model accuracy after unlearning in Section~\ref{sec6}, where our approach demonstrates substantial efficiency gains. In terms of storage, PDUDT requires $\mathcal{O}(k N_c d)$ memory, where $k$ is the number of training iterations and $N_c$ is the maximum number of neighbors of client $c$~\cite{PDUDT}.
This storage cost grows linearly with both the number of iterations and the network degree.
In sharp contrast, our approach requires only $\mathcal{O}(d)$ memory when using the Fisher approximation.
This dramatic reduction in space complexity makes our framework significantly better suited for resource-constrained decentralized environments.

\section{Theoretical Results}\label{sec5}
In this section, we establish the theoretical guarantees of our proposed framework.
Specifically, we prove that our method satisfies the formal $(\epsilon,\delta)$-certified unlearning definition and derive utility bounds showing that the unlearned model remains close to a model retrained from scratch.
Due to space constraints, we omit the detailed analysis of client-wise unlearning, which follows directly as a corollary of the sample-wise results. For clarity of exposition, we assume that all clients possess the same amount of local data, i.e., $n_1 = n_2 = \cdots = n_N = n$, and that each client issues an unlearning request of equal size $m$ when the optimization problem~\eqref{eq1} has been minimized after $K$ iterations.
Unless otherwise specified, $\|\cdot\|$ denotes the $\ell_2$ norm.
\subsection{Strongly Convex Loss}
We first consider the setting where the loss functions satisfy Assumption~\ref{assumptionStrong-convex}.

Before proving that Algorithm~\ref{alg} satisfies $(\epsilon,\delta)$-certified unlearning, we first present a series of lemmas that establish sensitivity upper bounds.
\begin{lemma}\label{lemmaUtility1}
For any client $i\in[N]$, let $x_{K,i}$ denote the local model obtained after $K$ training iterations using the full dataset, and $x_{K,i}^\prime$ the model obtained by retraining the learning algorithm after removing the deleted samples.
Then,
\begin{equation}\label{eq:utility1}
\|x_{K,i}^\prime - x_{K,i}\|
\le
\frac{2L m}{\lambda n}.
\end{equation}
\end{lemma}
\renewcommand{\IEEEQED}{\IEEEQEDopen}
\begin{proof}
Recall $f(x)=\frac{1}{Nn}\sum_{j=1}^N \sum_{l=1}^{n}F_j(x,\xi_{j,l})$. We define $\Tilde{f}^{S-U}(x)=\frac{1}{N(n-m)}\sum_{j=1}^N \sum_{l\in S_j\backslash U_j}F_j(x,\xi_{j,l})$, where $|U_j|=m$ for $\forall j\in[N]$. 
First we observe that 
\begin{align*}
    &Nn(f(x^\prime_{K,i})-f(x_{K,i}))\\
    =&N(n-m)(\Tilde{f}^{S-U}(x^\prime_{K,i})-\Tilde{f}^{S-U}(x_{K,i}))\\
    &+\sum_{j=1}^N \sum_{l\in U_j}(F_j(x^\prime_{K,i},\xi_{j,l})-F_j(x_{K,i},\xi_{j,l}))\\
    \le& NmL\|x^\prime_{K,i}-x_{K,i}\|.
\end{align*}
Then by the strong convexity of $f(x)$, we have 
\begin{align*}
    f(x^\prime_{K,i})-f(x_{K,i})\ge\frac{\lambda}{2}\|x^\prime_{K,i}-x_{K,i}\|^2.
\end{align*}
Thus we have 
\begin{align*}
    \frac{\lambda Nn}{2}\|x^\prime_{K,i}-x_{K,i}\|^2\le{NmL}\|x^\prime_{K,i}-x_{K,i}\|,
\end{align*}
which implies that $\|x^\prime_{K,i}-x_{K,i}\|\le\frac{2mL}{\lambda n}$.
\end{proof}

\begin{lemma}\label{lemmaUtility2}
    Let $\hat{x}_i$ denote the unlearned model obtained \emph{without} noise perturbation, then:
\begin{equation}\label{eq:Utility2}
    \|\hat{x}_{i}-x_{K,i}^\prime\|\leq \frac{2ML^{2}m^{2}}{\lambda^{3} n^{2}}.
\end{equation}
\end{lemma}
\renewcommand{\IEEEQED}{\IEEEQEDopen}
\begin{proof}
    Take Taylor's expansion for $\nabla\Tilde{f}^{S-U}(x_{K,i}^\prime)$ around the point $x_{K,i}$, we have
  $  \|\nabla\Tilde{f}^{S-U}(x_{K,i}^\prime)-\nabla\Tilde{f}^{S-U}(x_{K,i})-\nabla^2\Tilde{f}^{S-U}(x_{K,i})[x_{K,i}^\prime-x_{K,i}]\|\le\frac{M}{2}\|x_{K,i}^\prime-x_{K,i}\|^2.$
    Since $\nabla\Tilde{f}^{S-U}(x_{K,i}^\prime)=0$, we further have
    \begin{align*}
        &\|\nabla\Tilde{f}^{S-U}(x_{K,i})+\nabla^2\Tilde{f}^{S-U}(x_{K,i})[x_{K,i}^\prime-x_{K,i}]\|\\
        &\le\frac{M}{2}\|x_{K,i}^\prime-x_{K,i}\|^2.
    \end{align*}
    Note that
    \begin{align*}
        &\nabla\Tilde{f}^{S-U}(x_{K,i})\\
        =&\frac{1}{N(n-m)}\sum_{j=1}^N \sum_{l\in S_j\backslash U_j}\nabla F_j(x_{K,i},\xi_{j,l})\\
        =&\frac{n}{n-m}\nabla f(x_{K,i})-\frac{1}{N(n-m)}\sum_{j=1}^N\sum_{l\in U_j}\nabla F_j(x_{K,i},\xi_{j,l})\\
        =&-\frac{1}{N(n-m)}\sum_{j=1}^N\sum_{l\in U_j}\nabla F_j(x_{K,i},\xi_{j,l}).
    \end{align*}
    Hence, we have 
    \begin{align*}
        &\left\|\nabla^2\Tilde{f}^{S-U}(x_{K,i})[x_{K,i}^\prime\!-\!x_{K,i}]\!-\!\frac{\sum\limits_{j=1}^N\sum\limits_{l\in U_j}\nabla F_j(x_{K,i},\xi_{j,l})}{N(n-m)}\right\|\\
        &\le\frac{M}{2}\|x_{K,i}^\prime-x_{K,i}\|^2.
    \end{align*}
    Define a vector $v$ such that $x_{K,i}^\prime=x_{K,i}+\frac{1}{N(n-m)}[\nabla^2\Tilde{f}^{S-U}(x_{K,i})]^{-1}\sum_{j=1}^N\sum_{l\in U_j}\nabla F_j(x_{K,i},\xi_{j,l})+v$ and plug it in the above, we have
    \begin{align*}
        \left\|\nabla^2\Tilde{f}^{S-U}(x_{K,i})v\right\|\le\frac{M}{2}\|x_{K,i}^\prime-x_{K,i}\|^2.
    \end{align*}
    Since $\Tilde{f}^{S-U}(\cdot)$ is $\lambda$-strongly convex, we have that $\left\|\nabla^2\Tilde{f}^{S-U}(x_{K,i})v\right\|\ge\lambda\|v\|$ for any vector $v$. Then we get $\|v\|\le\frac{M}{2\lambda}\|x_{K,i}^\prime-x_{K,i}\|^2$. Finally, applying Lemma~\ref{lemmaUtility1}, we get that $\|v\|\le\frac{2ML^2m^2}{\lambda^3n^2}$. 
\end{proof}
Lemma~\ref{lemmaUtility2} quantifies the deviation between the unlearned model prior to noise perturbation and the model retrained on the dataset with the deleted samples removed.
Notably, the bound matches the state-of-the-art results in~\cite{RememberWhatYouWanttoForget} and scales as $\mathcal{O}(m^2/n^2)$.
This scaling implies that adding Gaussian noise with variance $\sigma_i^2=\Theta(m^2/n^2)$ suffices to achieve $(\epsilon,\delta)$-certified unlearning, as formalized below.
\begin{theorem}\label{th-ex-guarantee}
The outputs of the learning algorithm $\mathcal{A}$ and the unlearning algorithm $\mathcal{G}$ satisfy $(\epsilon,\delta)$-certified unlearning.
\end{theorem}
\renewcommand{\IEEEQED}{\IEEEQEDopen}
\begin{proof} 
Let $x_{K,i}$ denote the local model of client $i$ at iteration $k$ when trained on the full dataset, and let $\widetilde{x}_i$ denote the output of Algorithm~\ref{alg} without fine-tuning.
Let $\hat{x}_i$ be the corresponding unlearned model before noise perturbation.

Similarly, let $x_{K,i}^\prime$ denote the local model obtained at iteration $k$ when training on the retained dataset, and let $\widetilde{x}_i^\prime$ denote the output of Algorithm~\ref{alg} when no deletion request is issued.
Let $\hat{x}_i^\prime$ be the noise-free counterpart of $\widetilde{x}_i^\prime$.
We have $\hat{x}_i^\prime = x_{K,i}^\prime$.

Based on (\ref{eq:Utility2}), we define 
\begin{equation}
    ||\hat{x}_{i}-x_{K,i}^\prime||\leq\frac{2ML^{2}m^{2}}{\lambda^{3} n^{2}}: =\Delta F_i.
\end{equation}
\par 
Also, we note that the points $\widetilde{x}_{i}$ and $\widetilde{x}_{i}^\prime$ are computed with the noise $\nu_i\sim\mathcal{N}(0,\sigma_i^2\mathbb{I}_d)$ where $\sigma_i=(\Delta F_{i}/\varepsilon)\cdot\sqrt{2\ln(1.25/\delta)}$. Specifically, we have $\widetilde{x}_{i}=\hat{x}_{i}+\frac{1}{N}\sum_{i\in [N]}\nu_i$ and $\widetilde{x}_{i}^\prime=\hat{x}_{i}^\prime+\frac{1}{N}\sum_{i\in [N]}\nu_i^\prime$. Then, following along the lines of the proof of the differential privacy guarantee for the Gaussian mechanism\cite{dp}, we get that for any set $W $,
\begin{align}
        &\Pr(\widetilde{x}_{i}\in W )\leq e^{\varepsilon}\Pr\bigl(\widetilde{x}_{i}^\prime \in W )+\delta,~\text{and}\notag\\
        &\Pr\bigl(\widetilde{x}_{i}^\prime \in W )\leq e^{\varepsilon}\Pr(\widetilde{x}_{i}\in W )+\delta,
\end{align}
giving the desired unlearning guarantee.
\par Moreover, any following fine-tuning process does not hurt the $(\epsilon,\delta)$-unlearning guarantee. This property is similar to $immunity$ $to$ $post-processing$ in differential privacy\cite{dp}. For our $(\epsilon,\delta)$-unlearning algorithm $\mathcal{G}$, let $g:\mathcal{X}\to \mathcal{X}$ be a deterministic function. Fix any pair of neighboring databases $ a, b $ with $ \|a-b\|_{1} \leq 1 $, and fix any event $ \mathcal{W} \subseteq \mathcal{X} $. Let $\mathcal{T}=\{r \in \mathcal{X}: g(r) \in \mathcal{W}\}$. We then have:
\begin{align}
&\Pr\bigl[g(\mathcal{G}(a)) \in \mathcal{W}]  =\Pr\bigl[\mathcal{G}(a) \in \mathcal{T}] \notag\\\leq~& e^\epsilon \Pr\bigl[\mathcal{G}(b) \in \mathcal{T}]+\delta=e^\epsilon \Pr\bigl[g(\mathcal{G}(b)) \in \mathcal{W}]+\delta,
\end{align}
which is what we want.
\end{proof}

\begin{theorem}\label{th-ex-convergence}
    For any set $U_{i}\subseteq S_i$ of $m$ delete requests, the average unlearned model $\widetilde{x}:=\frac{1}{N}\sum_{i=1}^N\widetilde{x}_{i}$ satisfies:
\begin{align}
    &\mathbb{E}\left[f(\widetilde{x})-\min_{x\in\mathcal{X}}f(x)\right]\notag\\
    \leq &O\left(\frac{mL^2}{(n-m)\lambda } +\frac{\sqrt{d\ln(1/\delta)}ML^3m^2}{\lambda^3n^2\epsilon}\right).
\end{align}
\end{theorem}
\renewcommand{\IEEEQED}{\IEEEQEDopen}
\begin{proof}
We define $x^{*} \in \arg\min_{x \in \mathcal{X}} f(x)$. Then we have
\begin{align}\label{eq:th2proof}
    &\mathbb{E}[f(\widetilde{x})-f(x^*)]
\nonumber\\
    =&\mathbb{E}[f(\widetilde{x})-f(\bar{x})]
\nonumber\\
=& \mathbb{E}\left[f\left(\frac{1}{N}\sum_{i=1}^N\widetilde{x}_{i}\right)-f\left(\frac{1}{N}\sum_{i=1}^Nx_{K,i}\right)\right]
\nonumber\\
    \leq& L\cdot \mathbb{E}\left\| \frac{1}{N}\sum_{i=1}^N(\widetilde{x}_{i}-x_{K,i})\right\| 
\nonumber\\
\leq& \frac{L}{N} \sum_{i=1}^N\mathbb{E}\|\widetilde{x}_{i}-x_{K,i}\|.
\end{align}
\par
Also, we have that
\begin{align}\label{eq:widetilde_x-x_k}
    &\mathbb{E}[\| \widetilde{x}_{i}-x_{K,i} \| ]
\nonumber\\
    =&\mathbb{E} \| \frac{1}{N}\sum_{i=1}^{N}(\frac{1}{n-m}(\hat{H_i})^{-1}\sum_{\xi_i \in U_i}\nabla F_i(x_{K,i},\xi_i)+\nu_i ) \|
\nonumber\\
    \leq&\frac{1}{N} \sum_{i=1}^N\left(\mathbb{E} \| \frac{1}{n-m}(\hat{H_i})^{-1}\sum_{\xi_i \in U_i}\nabla F_i(x_{K,i},\xi_i)   \|+\mathbb{E} \left \| \nu_i  \right \|\right)
\nonumber\\
    \leq& \frac{1}{N(n\!-\!m)\lambda }\sum_{i=1}^N\sum_{\xi_i\in U_i}\mathbb{E}\left \| \nabla F_i(x_{K,i},\xi_i) \right \|\!+\!\frac{1}{N}\sum_{i=1}^N\sqrt{\mathbb{E}\left \| \nu_i\right \| ^2   }
\nonumber\\
    \leq& \frac{mL}{(n-m)\lambda } +\frac{\sqrt{d\ln(1/\delta)}2ML^2m^2}{\lambda^3n^2\epsilon}.
\end{align}
where the second-to-the-last inequality holds by the $L$-Lipschitzness of $F_i$ and an application of Jensen inequality to bound $\mathbb{E}[\left \| \nu_i\right \|]$. 
By substituting (\ref{eq:widetilde_x-x_k}) into (\ref{eq:th2proof}), we complete the proof.
\end{proof}
\subsection{General Convex Loss} 
In this subsection, we further consider the convex loss settings. The following lemma gives us a bound on the domain of interest, and thus allows us to bound the Lipschitz constant for loss function over this domain. Specifically, we can conclude that the regularized loss function $\widetilde{F}_i(x_i, \xi)$ is $\lambda$-strongly convex, $(\mu+\lambda)$-smooth, $(L+\lambda||x_i||)$-Lipschitz, and $M$-Hessian Lipschitz by Lemma \ref{lemmaDomain}, which makes it easy to extend our unlearning algorithm to general convex loss. 
\begin{assumption}
For any $\xi_i\in S_i$, all functions $F_i(x,\xi)$ are convex, $\mu$-smooth, $L$-Lipschitz and $M$-Hessian Lipschitz.
\end{assumption}
\par Our algorithms for convex losses are based on the strongly convex setting. Given the convex loss function $F_i(\cdot, \xi)$, we define the function $\widetilde{F}_i(\cdot, \xi)$ as $\widetilde{F}_i(x_i, \xi)=F_i(x_i, \xi)+\frac{\lambda}{2}\|x_i\|^{2}.$
\begin{lemma}\label{lemmaDomain}
Let $\widetilde{F}_i$ be defined as $\widetilde{F}_i(x_i, \xi)=F_i(x_i, \xi)+\frac{\lambda}{2}\|x_i\|^{2}$. Given a dataset $S_i$, let $\widetilde{x}^*_{i}$ denote the empirical risk minimizer of $\widetilde{F}_i$ on $S_i$. Then, the point $\widetilde{x}^*_{i}$ satisfies $\|\widetilde{x}^*_{i}\| \leq \frac{L}{\lambda}$.
\end{lemma}
\renewcommand{\IEEEQED}{\IEEEQEDopen}
\begin{proof}
By plugging in the definition of the function $\widetilde{F}_i$, we have $\frac{1}{n_i} \sum _{\xi_i \in S_i}\nabla \widetilde{F}_i(\widetilde{x}^*_{i},\xi_i)=\frac{1}{n_i} \sum _{\xi_i \in S_i}\nabla F_i(\widetilde{x}^*_{i},\xi_i)+\lambda {x}^*_{i}=0.$
 From an application of the Triangle inequality and the $L$-Lipschitzness of $F_i$, we can easily get that $\|\lambda \widetilde{x}^*_{i}\| \leq  \frac{1}{n_i} \sum _{\xi_i \in S_i}\| \nabla {F_i}(\widetilde{x}^*_{i},\xi_i)\| \leq L$. 
Thus we have $\|\widetilde{x}^*_{i}\| \leq \frac{L}{\lambda}$. Finally, we can bound the Lipschitz constant for loss function over this domain.
\end{proof}

\section{Experiments}\label{sec6}
\subsection{Experimental Setup}
We report experiments on several widely-used benchmark datasets, including CIFAR-10\cite{CIFAR-10}, MNIST\cite{MNIST} and Fashion-MNIST \cite{FMNIST}. We use a logistic regression model for MNIST dataset, considering a convex classification problem. For CIFAR-10 datasets, we use the ResNet-18 model\cite{resnet}. For Fashion-MNIST dataset, we use the CNN model. To show the advantages of our proposed algorithm, we compare it with two existing baselines: \textbf{RT}, which means retraining from scratch with the retained data or clients by D-PSGD\cite{dpsgd} and PDUDT\cite{PDUDT}. For all unlearning types, we employ \textbf{RT} as the primary baseline. Since PDUDT is designed specifically for client-wise unlearning, we use it exclusively in the client-wise scenario\cite{PDUDT}. Given that our work focuses on decentralized approximate unlearning, methods designed for either centralized or exact unlearning fall outside the scope of our comparative analysis\cite{hdus, FAT, Fedrecovery}.
\par Throughout our experiments, we consider a decentralized system with 10 clients, where the connections between clients are generated by the topologies including Ring graph and Erdős–Rényi random graph. Each client trains with a batch size of 100 for 1 epoch per round, with a learning rate of 0.001. The unlearning request is set to occur at round $K$, which is adjusted for each dataset based on its convergence rate. We also set the number of retraining rounds to be equal to $K$. In a non-IID setting, data across clients follow a Dirichlet distribution with $\alpha = 0.3$. 

\par
\begin{figure*}[t] 
    \centering
    \begin{subfigure}[b]{0.3\textwidth} 
        \includegraphics[width=\textwidth]{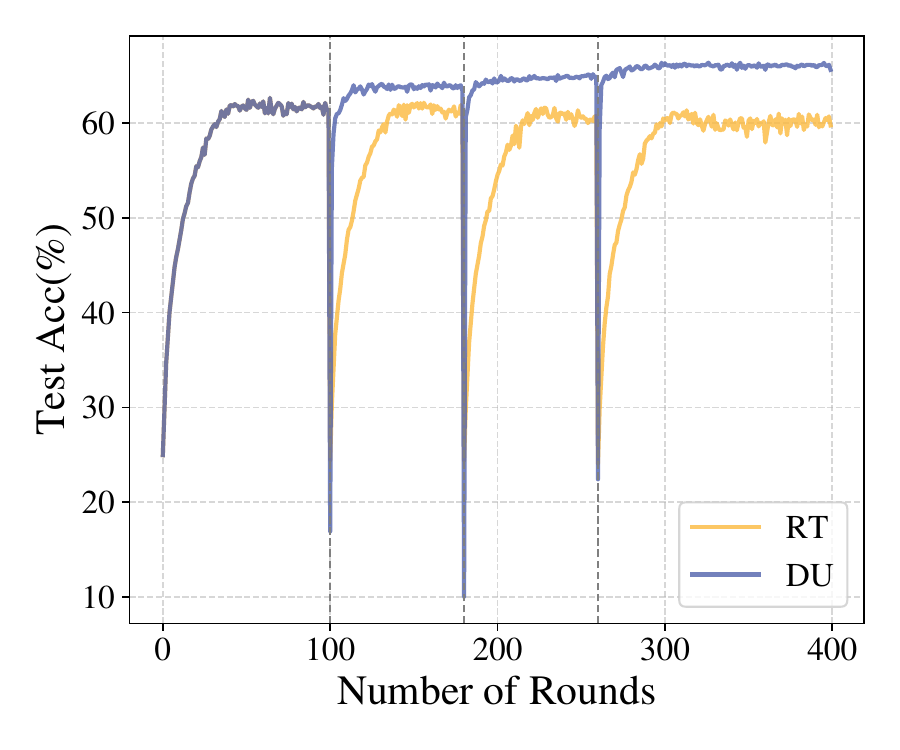}
        \caption{CIFAR-10}
        \label{fig:cifar10}
    \end{subfigure}
    \hfill 
    \begin{subfigure}[b]{0.3\textwidth}
        \includegraphics[width=\textwidth]{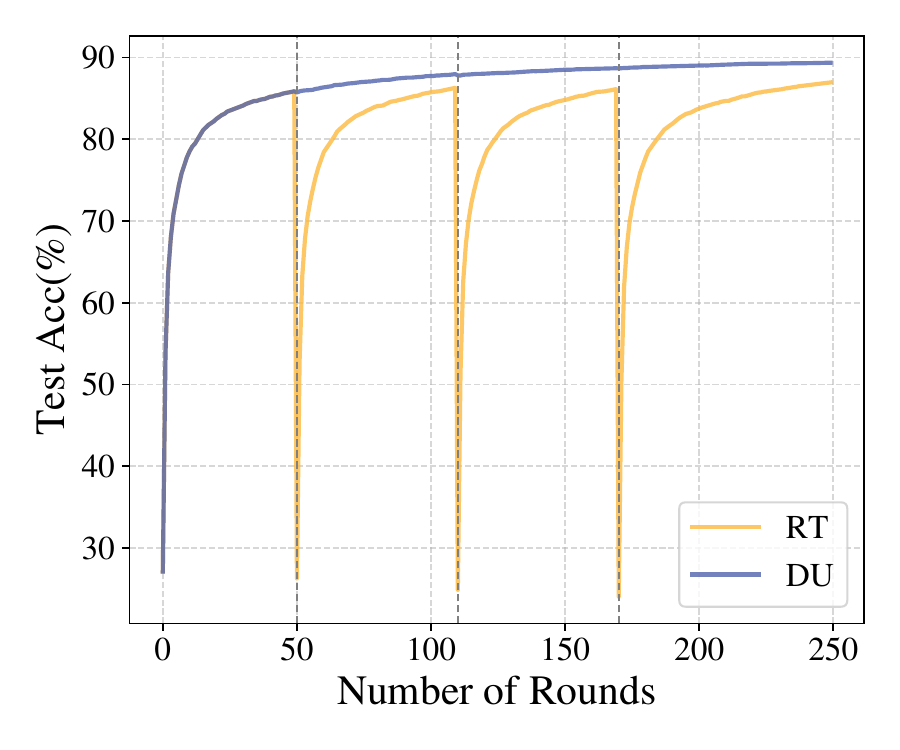}
        \caption{MNIST}
        \label{fig:mnist}
    \end{subfigure}
    \hfill
    \begin{subfigure}[b]{0.3\textwidth}
        \includegraphics[width=\textwidth]{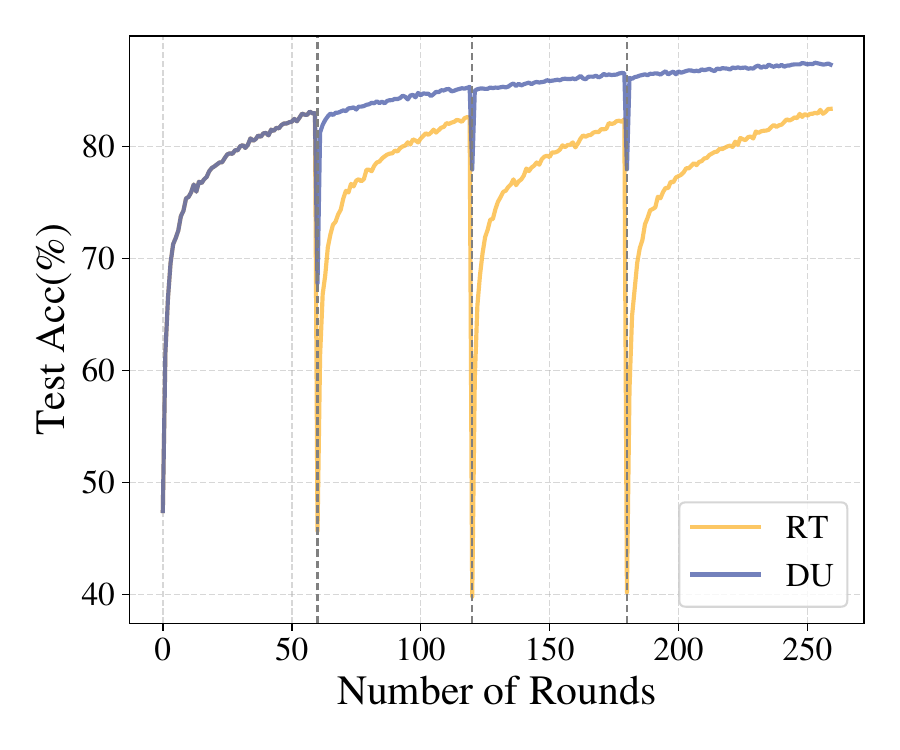}
        \caption{F-MNIST}
        \label{fig:F-MNIST}
    \end{subfigure}

    \vspace{0.1cm}
    \begin{subfigure}[b]{0.3\textwidth} 
        \includegraphics[width=\textwidth]{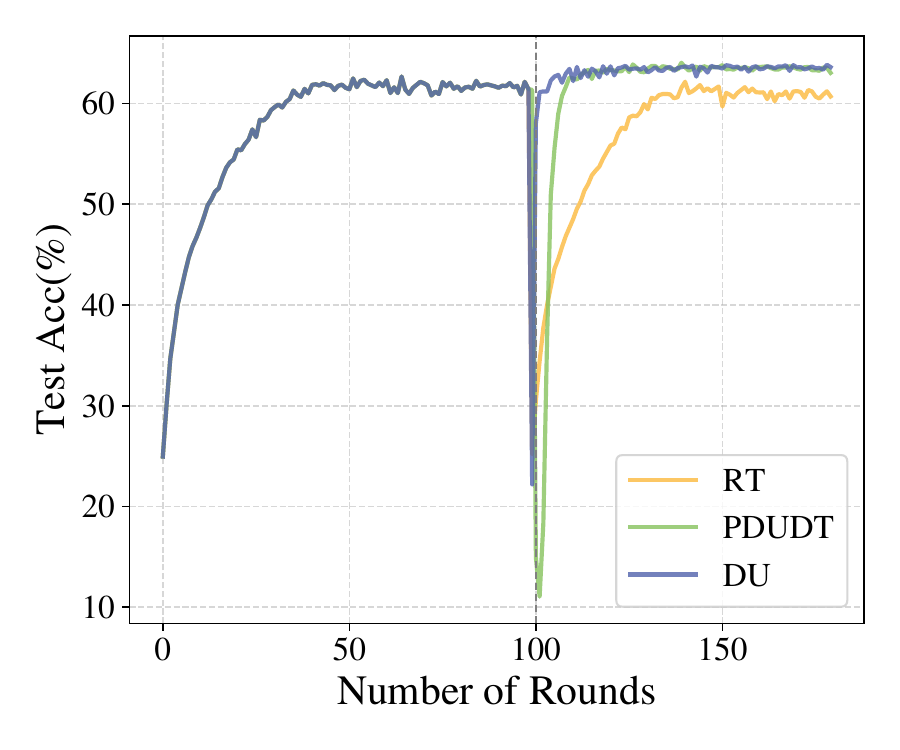}
        \caption{CIFAR-10}
        \label{fig:cifar10}
    \end{subfigure}
    \hfill 
    \begin{subfigure}[b]{0.3\textwidth}
        \includegraphics[width=\textwidth]{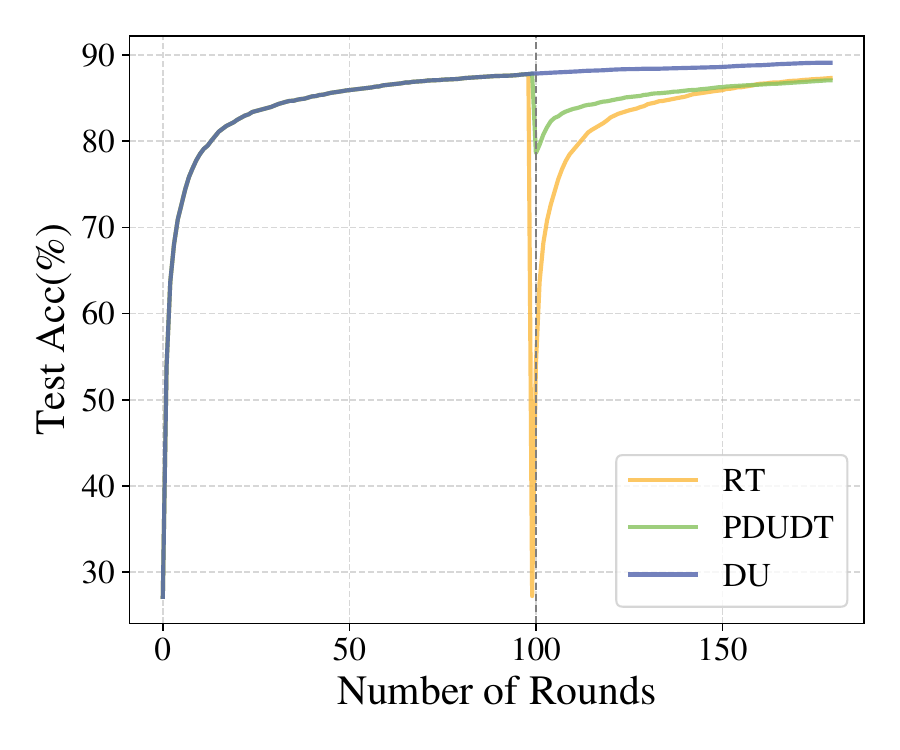}
        \caption{MNIST}
        \label{fig:mnist}
    \end{subfigure}
    \hfill
    \begin{subfigure}[b]{0.3\textwidth}
        \includegraphics[width=\textwidth]{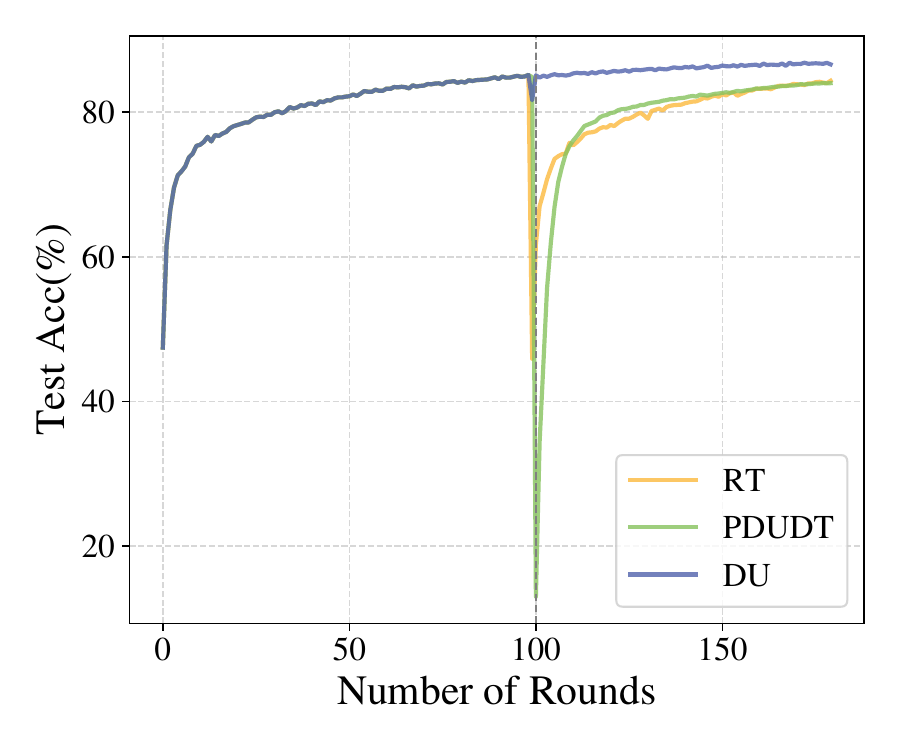}
        \caption{F-MNIST}
        \label{fig:F-MNIST}
    \end{subfigure}
    \caption{Comparison of the test accuracy of different methods and their changes after different unlearning requests. (a)-(c): sample-wise unlearning. (d)-(f): client-wise unlearning.}
    \label{fig1}
    \vspace{-1em}
\end{figure*}

 The unlearning tasks are structured as follows. For sample-wise unlearning in Table \ref{table-S-iid}-\ref{table-S-noniid}, we randomly select 10\% of each client’s data to be forgotten. For sample-wise unlearning in Fig.~\ref{fig1}, each client issues three sequential unlearning requests, cumulatively removing 20\% of its local data. For class-wise unlearning, we remove samples of class $0$ for all clients. For client-wise unlearning, we randomly remove a single client. For brevity, we use \textbf{ER} for the Erdős–Rényi graph, \textbf{DU} for our proposed unlearning algorithm, \textbf{Acc} for accuracy and \textbf{F-MNIST} for Fashion-MNIST.


\begin{table}
\footnotesize
\caption{Sample-wise unlearning in an IID setting}\label{table-S-iid}
\begin{center}
\begin{tabular}{ccccc}
\toprule
\textbf{Topology} & \textbf{Dataset} & \textbf{Method} & \textbf{\textit{Test Acc}} & \textbf{\textit{MIA Acc}} \\
\midrule
\multirow{6}{*}{RING} & \multicolumn{1}{c}{\multirow{2}{*}{CIFAR-10}} & RT & $\textbf{58.99}_{\pm 0.40\%}$ & $\textbf{50.25}_{\pm 0.98\%}$  \\    
                      & \multicolumn{1}{c}{} & DU & $55.28_{\pm 0.79\%}$ & $50.72_{\pm 1.48\%}$  \\ 
                      & \multicolumn{1}{c}{\multirow{2}{*}{MNIST}} & RT & $86.34_{\pm 0.23\%}$ & $52.29_{\pm 0.27\%}$  \\ 
                      & \multicolumn{1}{c}{} & DU & $\textbf{86.63}_{\pm 0.04\%}$ & $\textbf{51.96}_{\pm 0.53\%}$  \\ 
                      & \multicolumn{1}{c}{\multirow{2}{*}{F-MNIST}} & RT & $84.24_{\pm 0.16\%}$ & $49.53_{\pm 0.19\%}$  \\ 
                      & \multicolumn{1}{c}{} & DU & $\textbf{84.40}_{\pm 0.23\%}$ & $\textbf{50.00}_{\pm 0.58\%}$  \\
\midrule
\multirow{6}{*}{ER} & \multicolumn{1}{c}{\multirow{2}{*}{CIFAR-10}} & RT & $\textbf{62.20}_{\pm 0.46\%}$ & $\textbf{50.48}_{\pm 0.67\%}$  \\    
                      & \multicolumn{1}{c}{} & DU & $57.82_{\pm 0.23\%}$ & $51.47_{\pm 0.70\%}$  \\ 
                      & \multicolumn{1}{c}{\multirow{2}{*}{MNIST}} & RT & $86.25_{\pm 0.10\%}$ & $51.60_{\pm 0.56\%}$  \\ 
                      & \multicolumn{1}{c}{} & DU & $\textbf{86.63}_{\pm 0.04\%}$ & $\textbf{51.51}_{\pm 0.42\%}$  \\ 
                      & \multicolumn{1}{c}{\multirow{2}{*}{F-MNIST}} & RT & $84.21_{\pm 0.15\%}$ & $50.53_{\pm 0.72\%}$  \\ 
                      & \multicolumn{1}{c}{} & DU & $\textbf{84.42}_{\pm 0.20\%}$ & $\textbf{50.42}_{\pm 0.83\%}$  \\
\bottomrule
\end{tabular}
\end{center}
\vspace{-1em}
\end{table}

\begin{table}
\footnotesize
\caption{Sample-wise unlearning in a non-IID setting}\label{table-S-noniid}
\begin{center}
\begin{tabular}{ccccc}
\toprule
\textbf{Topology} & \textbf{Dataset} & \textbf{Method} & \textbf{\textit{Test Acc}} & \textbf{\textit{MIA Acc}} \\
\midrule
\multirow{6}{*}{RING} & \multicolumn{1}{c}{\multirow{2}{*}{CIFAR-10}} & RT & $\textbf{51.68}_{\pm 2.23\%}$ & $50.40_{\pm 0.37\%}$  \\    
                      & \multicolumn{1}{c}{} & DU & $45.06_{\pm 3.74\%}$ & $\textbf{50.20}_{\pm 1.29\%}$  \\ 
                      & \multicolumn{1}{c}{\multirow{2}{*}{MNIST}} & RT & $86.59_{\pm 0.13\%}$ & $\textbf{50.27}_{\pm 0.66\%}$  \\ 
                      & \multicolumn{1}{c}{} & DU & $\textbf{86.95}_{\pm 0.15\%}$ & $50.69_{\pm 0.17\%}$  \\ 
                      & \multicolumn{1}{c}{\multirow{2}{*}{F-MNIST}} & RT & $\textbf{82.41}_{\pm 1.12\%}$ & $\textbf{50.00}_{\pm 0.81\%}$  \\ 
                      & \multicolumn{1}{c}{} & DU & $81.14_{\pm 1.62\%}$ & $49.98_{\pm 0.17\%}$  \\
\midrule
\multirow{6}{*}{ER} & \multicolumn{1}{c}{\multirow{2}{*}{CIFAR-10}} & RT & $\textbf{54.95}_{\pm 1.34\%}$ & $\textbf{49.70}_{\pm 0.75\%}$  \\    
                      & \multicolumn{1}{c}{} & DU & $47.53_{\pm 3.40\%}$ & $50.77_{\pm 0.70\%}$  \\ 
                      & \multicolumn{1}{c}{\multirow{2}{*}{MNIST}} & RT & $84.16_{\pm 4.22\%}$ & $\textbf{52.56}_{\pm 1.55\%}$  \\ 
                      & \multicolumn{1}{c}{} & DU & $\textbf{87.21}_{\pm 0.33\%}$ & $52.79_{\pm 1.75\%}$  \\ 
                      & \multicolumn{1}{c}{\multirow{2}{*}{F-MNIST}} & RT & $\textbf{82.35}_{\pm 1.22\%}$ & $49.50_{\pm 0.66\%}$  \\ 
                      & \multicolumn{1}{c}{} & DU & $81.20_{\pm 1.40\%}$ & $\textbf{49.78}_{\pm 0.88\%}$  \\
\bottomrule
\end{tabular}
\end{center}
\vspace{-1em}
\end{table}

\begin{table}
\footnotesize
\caption{Class-wise unlearning in an IID setting}\label{table-C-iid}
\begin{center}
\begin{tabular}{ccccc}
\toprule
\textbf{Topology} & \textbf{Dataset} & \textbf{Method} & \textbf{\textit{Test Acc}} & \textbf{\textit{MIA Acc}} \\
\midrule
\multirow{6}{*}{RING} & \multicolumn{1}{c}{\multirow{2}{*}{CIFAR-10}} & RT & $\textbf{54.67}_{\pm 0.50\%}$ & $\textbf{51.08}_{\pm 3.27\%}$  \\    
                      & \multicolumn{1}{c}{} & DU & $53.35_{\pm 1.07\%}$ & $53.50_{\pm 2.35\%}$  \\ 
                      & \multicolumn{1}{c}{\multirow{2}{*}{MNIST}} & RT & $77.71_{\pm 0.04\%}$ & $53.66_{\pm 2.25\%}$  \\ 
                      & \multicolumn{1}{c}{} & DU & $\textbf{84.01}_{\pm 0.12\%}$ & $\textbf{52.55}_{\pm 0.36\%}$  \\ 
                      & \multicolumn{1}{c}{\multirow{2}{*}{F-MNIST}} & RT & $77.88_{\pm 0.13\%}$ & $\textbf{49.58}_{\pm 4.23\%}$  \\ 
                      & \multicolumn{1}{c}{} & DU & $\textbf{78.02}_{\pm 0.14\%}$ & $49.42_{\pm 3.83\%}$  \\
\midrule
\multirow{6}{*}{ER} & \multicolumn{1}{c}{\multirow{2}{*}{CIFAR-10}} & RT & $\textbf{57.77}_{\pm 0.64\%}$ & $\textbf{49.75}_{\pm 3.02\%}$  \\    
                      & \multicolumn{1}{c}{} & DU & $55.20_{\pm 1.02\%}$ & $50.40_{\pm 0.44\%}$  \\ 
                      & \multicolumn{1}{c}{\multirow{2}{*}{MNIST}} & RT & $77.71_{\pm 0.04\%}$ & $53.66_{\pm 2.25\%}$  \\ 
                      & \multicolumn{1}{c}{} & DU & $\textbf{84.00}_{\pm 0.14\%}$ & $\textbf{52.55}_{\pm 0.42\%}$  \\ 
                      & \multicolumn{1}{c}{\multirow{2}{*}{F-MNIST}} & RT & $77.86_{\pm 0.13\%}$ & $\textbf{49.42}_{\pm 3.80\%}$  \\ 
                      & \multicolumn{1}{c}{} & DU & $\textbf{78.04}_{\pm 0.15\%}$ & $49.33_{\pm 3.95\%}$  \\
\bottomrule
\end{tabular}
\end{center}
\vspace{-1em}
\end{table}

\begin{table}
\footnotesize
\caption{Class-wise unlearning in a non-IID setting}\label{table-C-noniid}
\begin{center}
\begin{tabular}{ccccc}
\toprule
\textbf{Topology} & \textbf{Dataset} & \textbf{Method} & \textbf{\textit{Test Acc}} & \textbf{\textit{MIA Acc}} \\
\midrule
\multirow{6}{*}{RING} & \multicolumn{1}{c}{\multirow{2}{*}{CIFAR-10}} & RT & $\textbf{48.07}_{\pm 2.84\%}$ & $51.83_{\pm 2.18\%}$  \\    
                      & \multicolumn{1}{c}{} & DU & $43.15_{\pm 2.72\%}$ & $\textbf{51.25}_{\pm 1.87\%}$  \\ 
                      & \multicolumn{1}{c}{\multirow{2}{*}{MNIST}} & RT & $78.10_{\pm 0.14\%}$ & $\textbf{54.08}_{\pm 1.10\%}$  \\ 
                      & \multicolumn{1}{c}{} & DU & $\textbf{86.31}_{\pm 0.43\%}$ & $54.34_{\pm 0.36\%}$  \\ 
                      & \multicolumn{1}{c}{\multirow{2}{*}{F-MNIST}} & RT & $\textbf{76.26}_{\pm 1.09\%}$ & $\textbf{50.42}_{\pm 0.92\%}$  \\ 
                      & \multicolumn{1}{c}{} & DU & $76.22_{\pm 0.90\%}$ & $50.92_{\pm 2.29\%}$  \\
\midrule
\multirow{6}{*}{ER} & \multicolumn{1}{c}{\multirow{2}{*}{CIFAR-10}} & RT & $\textbf{50.92}_{\pm 1.54\%}$ & $\textbf{49.67}_{\pm 0.96\%}$  \\    
                      & \multicolumn{1}{c}{} & DU & $46.16_{\pm 2.99\%}$ & $54.58_{\pm 0.42\%}$  \\ 
                      & \multicolumn{1}{c}{\multirow{2}{*}{MNIST}} & RT & $81.34_{\pm 3.83\%}$ & $53.59_{\pm 1.89\%}$  \\ 
                      & \multicolumn{1}{c}{} & DU & $\textbf{86.79}_{\pm 0.15\%}$ & $\textbf{53.02}_{\pm 1.54\%}$  \\ 
                      & \multicolumn{1}{c}{\multirow{2}{*}{F-MNIST}} & RT & $76.00_{\pm 1.16\%}$ & $51.25_{\pm 0.00\%}$  \\ 
                      & \multicolumn{1}{c}{} & DU & $\textbf{76.16}_{\pm 0.88\%}$ & $\textbf{51.00}_{\pm 2.47\%}$  \\
\bottomrule
\end{tabular}
\end{center}
\vspace{-1em}
\end{table}

\begin{figure*}[t] 
    \centering
    \begin{subfigure}[b]{0.3\textwidth} 
        \includegraphics[width=\textwidth]{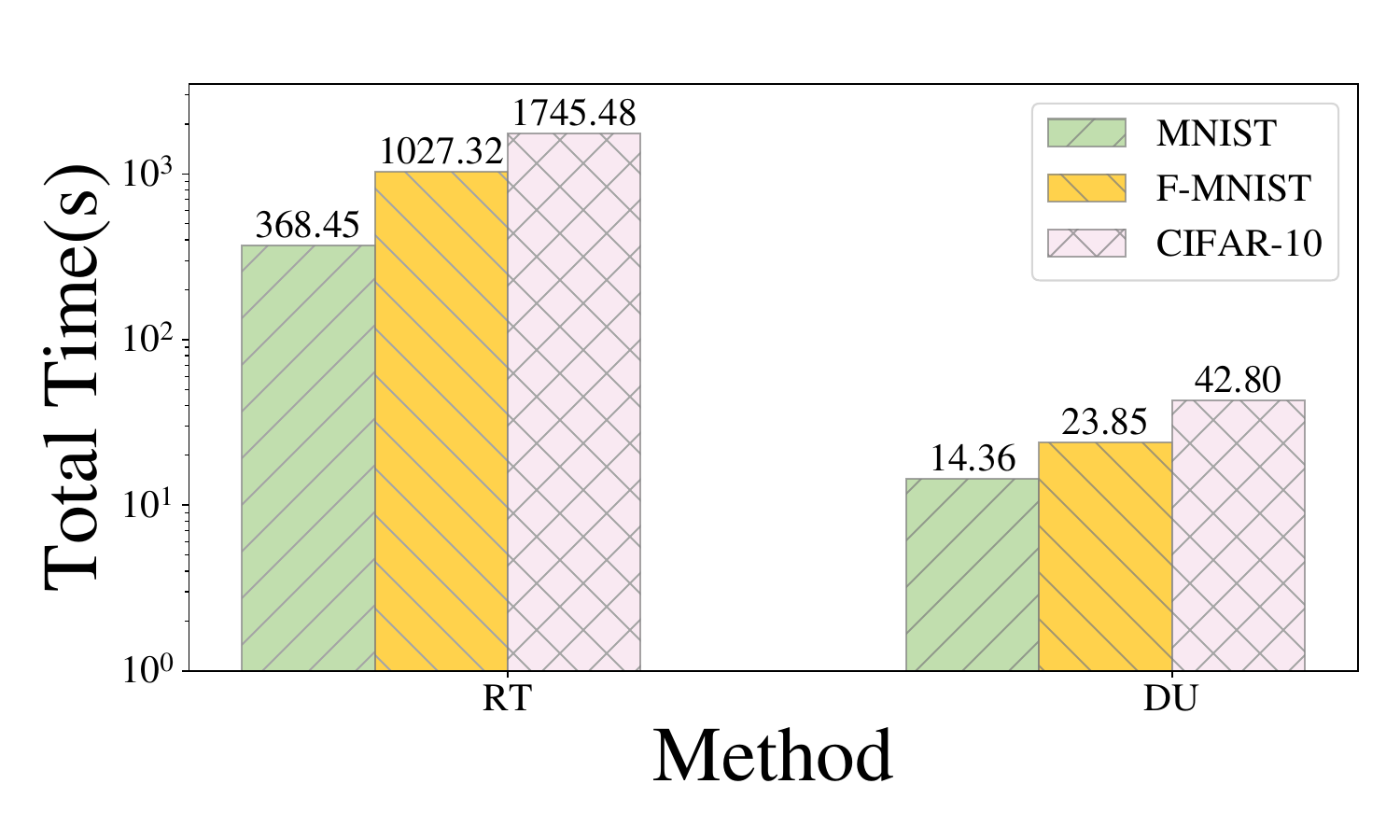}
        \caption{Sample-wise}
        \label{fig:time_sample}
    \end{subfigure}
    \hfill 
    \begin{subfigure}[b]{0.3\textwidth}
        \includegraphics[width=\textwidth]{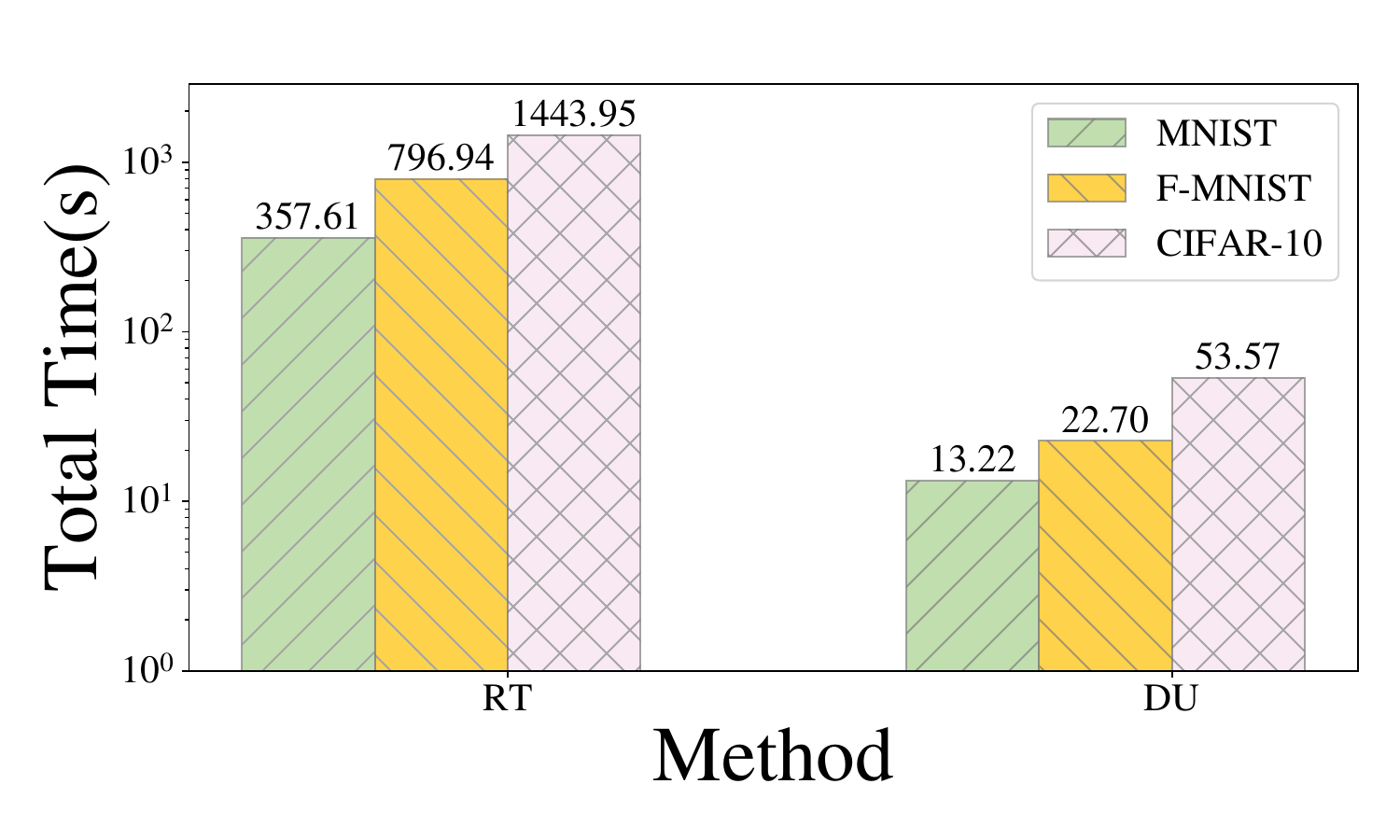}
        \caption{Class-wise}
        \label{fig:time_class}
    \end{subfigure}
    \hfill
    \begin{subfigure}[b]{0.3\textwidth}
        \includegraphics[width=\textwidth]{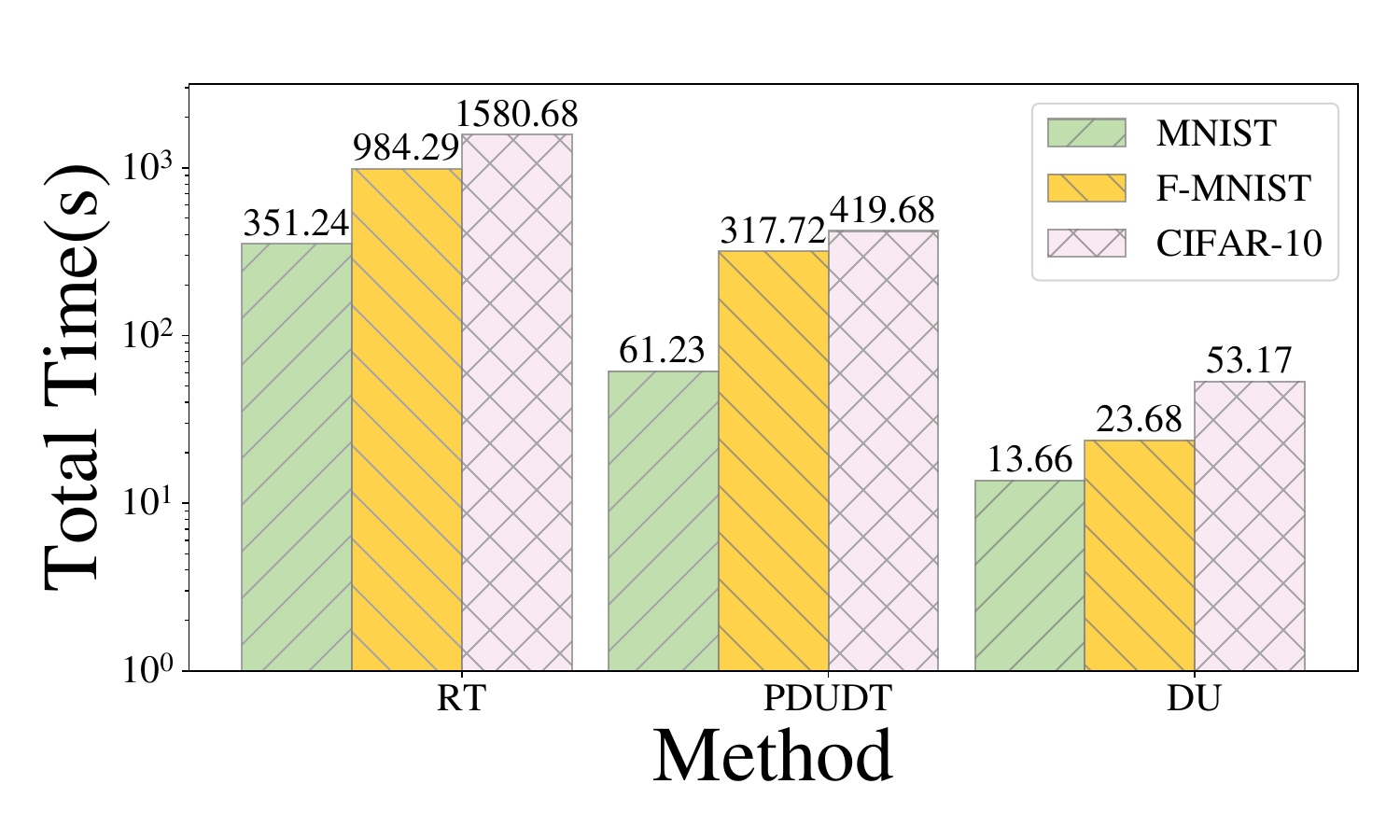}
        \caption{Client-wise}
        \label{fig:time_client}
    \end{subfigure}
    \caption{Comparison of the computation time of different methods.}
    \label{fig2}
    \vspace{-1em}
\end{figure*}

\subsection{Experimental Results}
\subsubsection{Unlearning Performance}
We measure and report the best test accuracy across a comprehensive set of scenarios. As shown in Tables \ref{table-S-iid}-\ref{table-C-noniid}, our unlearning algorithm consistently demonstrates superior performance in all settings including different data distributions and topologies. It successfully preserves model utility, delivering performance comparable to fully retraining models with substantial time savings. 
\par Furthermore, we conduct additional experiments for IID setting with Erdős–Rényi random graph. We primarily focus on sample-wise and client-wise unlearning, with the corresponding results shown in Fig.~\ref{fig1}. We measure and present the model test accuracy throughout the whole process including the initial training, issuing unlearning requests, and the recovery from data deletion using different methods. Our algorithm's consistent performance across the sequential unlearning requests serves as a strong testament to its stability. It should be noted that, to maintain the consistency of the x-axis, the \textbf{DU} method performed multiple rounds of fine-tuning in the experiments for Fig.~\ref{fig1}, whereas only a single round was executed in all other experiments. In fact, our algorithm requires just one round of fine-tuning to recover accuracy. Therefore, the curve of \textbf{DU} exhibits a sharp rise after each deletion request, while \textbf{RT} and PDUDT require multiple rounds to regain comparable performance. Last, on simpler datasets such as MNIST, our algorithm exhibits negligible performance degradation after unlearning requests. This is because the corrective model updates alone are sufficient to maintain model accuracy, which further demonstrates the effectiveness of our proposed method.
\subsubsection{Membership Inference Attack}
We employ membership inference attacks (MIA) to assess the privacy leakage risk of the unlearning process. Specifically, MIA aims to infer whether a given data sample was part of the model's training set. It quantifies the extend to which the forgotten data remains identifiable within the unlearned model. Consequently, an MIA accuracy approaching 50\% (which is equivalent to a random guess) signifies a more effective and secure unlearning process. As reported in Tables \ref{table-S-iid}-\ref{table-C-noniid}, the MIA accuracy of our method consistently approaches the random guess. This indicates that an attacker cannot reliably infer training set membership, thus confirming the high efficacy of our algorithm. Additionally, the overall performance of \textbf{DU} remains on par with \textbf{RT} method.
\subsubsection{Unlearning Efficiency}
To evaluate the efficiency of each method, we measure and present the time elapsed from the issuance of an unlearning request until the model's performance stabilizes in IID setting with Erdős–Rényi random graph. To ensure a fair and effective comparison, the recorded time for \textbf{DU} encompasses the entire unlearning and fine-tuning process. The experimental results in Fig.~\ref{fig2} demonstrate that our algorithm substantially reduces the unlearning time. Compared with \textbf{RT} method, we achieve approximately 97\% time savings across all datasets and unlearning types. We note that the numerical values shown in Fig.~\ref{fig2} represent actual time, whereas the bar chart is plotted using a logarithmic scale.  






\section{Conclusion}
In this paper, we proposed a certified decentralized unlearning framework based on Newton-style updates coupled with efficient Hessian approximations.
Our approach achieves high efficiency in both computational cost and storage overhead.
We theoretically proved that the proposed algorithm satisfies the formal $(\epsilon,\delta)$-certified unlearning guarantee and established rigorous utility bounds showing that the unlearned model remains close to a fully retrained one.
Extensive experimental results further validate the effectiveness and efficiency of our framework. While our analysis focuses on convex objectives, extending certified decentralized unlearning to non-convex settings remains an important and challenging direction for future work.

\section*{Acknowledgment}

Youming Tao was supported by the National Science Foundation of China (NSFC) under Grant 623B2068. Shuzhen Chen was supported in part by the Natural Science Foundation of Shandong Province under Grant No. ZR2025QC1537, in part by the Postdoctoral Fellowship Program of the China Postdoctoral Science Foundation (CPSF) under Grant No. GZC20251040, in part by the Fundamental Research Funds for the Central Universities under Grant No. 202513024, and in part by the Postdoctoral Program of Qingdao under Grant No. QDBSH20250102012. Dongxiao Yu was supported in part by the Major Basic Research Program of the Shandong Provincial Natural Science Foundation under Grant No. ZR2025ZD18, and in part by the Joint Key Funds of the National Natural Science Foundation of China under Grant No. U23A20302.

\newpage
\bibliographystyle{unsrt}
\bibliography{references}

\end{document}